\documentclass[conference]{IEEEtran}
\IEEEoverridecommandlockouts
\usepackage{cite}
\usepackage{amsmath,amssymb,amsfonts}
\usepackage{algorithmic}
\usepackage{graphicx}
\usepackage{textcomp}
\usepackage{xcolor}
\def\BibTeX{{\rm B\kern-.05em{\sc i\kern-.025em b}\kern-.08em
    T\kern-.1667em\lower.7ex\hbox{E}\kern-.125emX}}
\usepackage{url}
\usepackage{multirow}
\usepackage{booktabs}
\usepackage{xspace}
\usepackage{balance}
\def\myBezier{B\'{e}zier\space}
\begin{document}

\title{SketchAnimator: Animate Sketch via Motion Customization of Text-to-Video Diffusion Models}

\author{\IEEEauthorblockN{Ruolin Yang\textsuperscript{1},
Da Li\textsuperscript{2},
Honggang Zhang\textsuperscript{1} and
Yi-Zhe Song\textsuperscript{2}}
\IEEEauthorblockA{\textsuperscript{1}PRIS, School of Artificial Intelligence,
Beijing University of Posts and Telecommunications, China}
\IEEEauthorblockA{\textsuperscript{2}SketchX, CVSSP, University of Surrey, United Kingdom}
}


\maketitle

\begin{abstract}

Sketching is a uniquely human tool for expressing ideas and creativity. The animation of sketches infuses life into these static drawings, opening a new dimension for designers. Animating sketches is a time-consuming process that demands professional skills and extensive experience, often proving daunting for amateurs. In this paper, we propose a novel sketch animation model \textit{SketchAnimator}, which enables adding \emph{creative} motion to a given sketch, like ``a jumping car''. Namely, given an input sketch and a reference video, we divide the sketch animation into three stages: Appearance Learning, Motion Learning and Video Prior Distillation. In stages 1 and 2, we utilize LoRA to integrate sketch appearance information and motion dynamics from the reference video into the pre-trained T2V model. In the third stage, we utilize Score Distillation Sampling (SDS) to update the parameters of the \myBezier curves in each sketch frame according to the acquired motion information. Consequently, our model produces a sketch video that not only retains the original appearance of the sketch but also mirrors the dynamic movements of the reference video. We compare our method with alternative approaches and demonstrate that it generates the desired sketch video under the challenge of one-shot motion customization.


\end{abstract}

\vspace{-0.25cm}
\section{Introduction}
Free-hand sketching is an effective tool for showcasing unique creativity through abstract expression. Dynamic sketches, in particular, have a broader range of applications (e.g., animated advertisements, educational videos and storyboarding). Users often express a desire to create imaginative concepts that defy real-world constraints using sketches, such as ``a running clock". Recently, personalized and customized video generation has been explored~\cite{zhao2023motiondirector,wei2023dreamvideo, ren2024customize}. The objective of this task is to adapt pre-trained foundational models to generate videos depicting a specific motion concept, utilizing reference videos as guidance. However, due to the significant disparity between sketches and the data used to train large-scale generative models, research on customized and personalized dynamic sketch videos is still relatively limited. Namely, directly applying large-scale pre-trained T2V diffusion models presents challenges in terms of generalization to sketch domains. In this work, given a user sketch and a reference video sequence depicting a moving object, our framework generates a video in which the sketch is animated according to the driving sequence without changing its appearance.

We observe three key issues in previous customized sketch animation methods: (i) rely on user annotations: Su \emph{et al.}\cite{su2018live} developed an interactive system where users input a motion-indicating video sequence and mark key points on sketches to generate dynamic animations. However, this method depends heavily on accurate marking and remains challenging and time-consuming; (ii) fail to decouple motion signals apart from appearance: traditional methods~\cite{siarohin2019first, siarohin2021motion, zhao2022thin, hong2022depth} focus on specific domains such as faces, human figures, and related subjects. They train on datasets that gather numerous similar actions and subjects, relying on specific key-points. This limits customization in video generation. Video quality declines when objects in the driving video and source image vary in domain or shape subtly~\cite{xu2022motion}; (iii) lack of creativity: current state-of-the-art method, live sketch \cite{gal2023breathing}, animates a single-subject sketch using motion priors from a large pre-trained T2V diffusion model. However, this model may struggle with specific motion concepts and novel subject-motion combinations unseen in real life. For example, generating a video of ``A jumping tree" could challenge producing realistic and consistent results aligned with the description.

\vspace{-0.1cm}
To mitigate these issues, we propose SketchAnimator, a novel approach for animating a user-provided sketch with a specific motion extracted from a reference video, \emph{i.e.} one-shot customized sketch video generation. To avoid human annotations and subtle shape problems, 
we employ Low-Rank Adaptations~\cite{hu2021lora} to fine-tune pre-trained T2V models on sketches and videos. This allows us to extract the motion information embedded within the video sequence and the subject-specific details inherent in the sketch in a staged manner.  
To enhance diversity and creativity, we combine identity features with motion patterns to generate a video prior capable of predicting the content of the target sketch animation. However, direct inference faces challenges in preserving both the structural integrity of the sketch and the fidelity of motion information. Therefore, building on prior research~\cite{vinker2022clipasso, vinker2023clipascene, xing2024diffsketcher}, we employ a differentiable Bézier curves rendering to achieve adaptable sketch representation. Additionally, we show how to leverage a score distillation sampling (SDS) loss~\cite{poole2022dreamfusion} to guide the sketch generation while staying aligned with the customized video prior. We show superior results compared with the traditional motion transfer method and advanced pixel-based diffusion generation models. Moreover, our work allows users to animate sketches with motions that are not present in the training set of large-scale pre-trained video diffusion models.
\vspace{-0.2cm}
\section{Method}
\label{sec:method}
In this section, we begin with preliminaries. Next, we outline the details of SketchAnimator, which aim to address three key challenges: 1) how to preserve the characteristics of the original sketch; 2) how to extract the motion pattern from the reference video; and 3) how to add motion information to the sketch without changing its appearance in a one-shot manner:

\vspace{-0.2cm}
\subsection{Preliminaries}
\noindent\textbf{Video Diffusion Models}
Similar to image diffusion models, video diffusion models (VDM)~\cite{wang2023modelscope, chen2023videocrafter1}
learn a distribution representing video data through denoising process. During training, an autoencoder map 
the video sample $x$ of length $F$ into a latent representation $z_0 \in \mathbf{R}^{F \times h \times w \times c}$. Next, the forward process gradually adds Gaussian noise $\epsilon \sim \mathcal{N}(0, \mathbf{\mathrm{I}})$ to the latent code to obtain a noised data, as shown as follows:
\begin{equation}
    z_t = \alpha_t z_0 + \sigma_t \epsilon,
\end{equation}
where $\alpha_t$, $\sigma_t$ are two predetermined schedule~\cite{dhariwal2021diffusion} of random time step $t\sim \mathcal{U}(0, \mathbf{\mathrm{T}})$.
Video
diffusion model is conditioned on text prompts $y$ and adopt a 3D U-Net $\epsilon_\theta$ to perform denoising. Let $\tau_\theta(y)$ be the text encoder. The following reweighted variational bound is employed for optimization:
\begin{equation}
    \mathcal{L} = \mathbf{E}_{z_0, y, t, \epsilon}\left[\left\| \epsilon - \epsilon_{W}\left(z_t, \tau_\theta(y), t\right)\right\|_2^2\right],
\label{eq:VDM_loss}
\end{equation}

\noindent\textbf{Low-Rank Adaptation}
Low-Rank Adaptation (LoRA)~\cite{hu2021lora} method enables
efficient adaptation of large pre-trained language models to
downstream tasks. It predicts $\Delta W$ to update the pre-trained weight matrix $W_0 \in \mathbf{R}^{d \times k}$. Specifically, $\Delta W$ are decomposed by weight $B\in \mathbf{R}^{d \times r}$ and $A\in \mathbf{R}^{r \times k}$. After fine-tuning, $\Delta W$ can be merged into $W_0$ as a plug-and-play module which altered the direction of the model's prediction as:
\begin{equation}
    W = W_{0} + \alpha \Delta W = W_{0} + \alpha B A,
\end{equation}
where $d$ is the input dimension and $k$ is the output dimension with $r\ll\min(d, k)$.

\begin{figure}[ht]
  \centering
  \setlength{\abovecaptionskip}{-3pt}
  \includegraphics[width=1.0\linewidth]{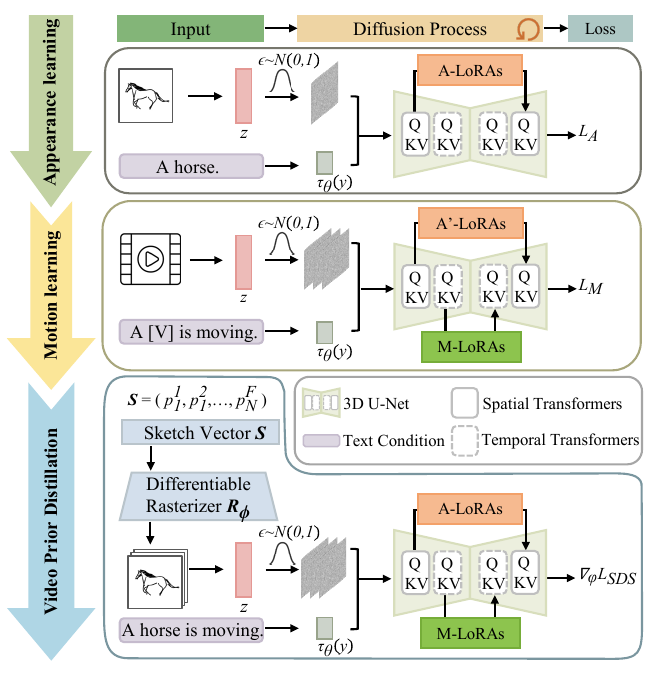}
  \caption{Illustration of the proposed SketchAnimator pipeline. In appearance learning stage, the trainable A-LoRAs are applied to absorb the visual information of sketch. In Motion learning stage, we separate motion information using A'-LoRAs and M-LoRAs to reconstruct the video. During Video Prior Distillation stage, we combine two LoRAs and update the parameters of the \myBezier curves in each sketch frame by SDS loss.}
  \label{fig:pipline}
\end{figure}
\noindent\textbf{Score-Distillation Sampling}
DreamFusion~\cite{poole2022dreamfusion} first proposed the score-distillation sampling (SDS) loss for optimizing pre-trained diffusion models to guided the 3D generation process.
Given an arbitrary differentiable parametric function that
renders images x (\emph{e.g.}, a NeRF), $g_\phi$, the gradient of the diffusion loss function with respect to the parameters  $\phi$ specified as SDS loss is given by:
\vspace{-0.1cm}
\begin{equation}\label{eq:sds_loss}
    \nabla_\phi \mathcal{L}_{SDS} = \left[ w(t)(\epsilon_{W}(z_t,\tau_\theta(y)),t ) \frac{\partial x}{\partial \phi} \right] ,
\end{equation}
where $w(t)$ is a weighting function. 

In our paper, sketch is rendering by a differentiable rasterizer~\cite{li2020differentiable} at stroke level following previous sketch generation works~\cite{xing2024diffsketcher, vinker2022clipasso, vinker2023clipascene}. Hence, we utilize SDS loss to generate sketch videos via
optimization from a customized diffusion model.

\begin{figure*}[t]
  \centering
  \setlength{\abovecaptionskip}{-3pt}
  \includegraphics[width=1.0\linewidth]{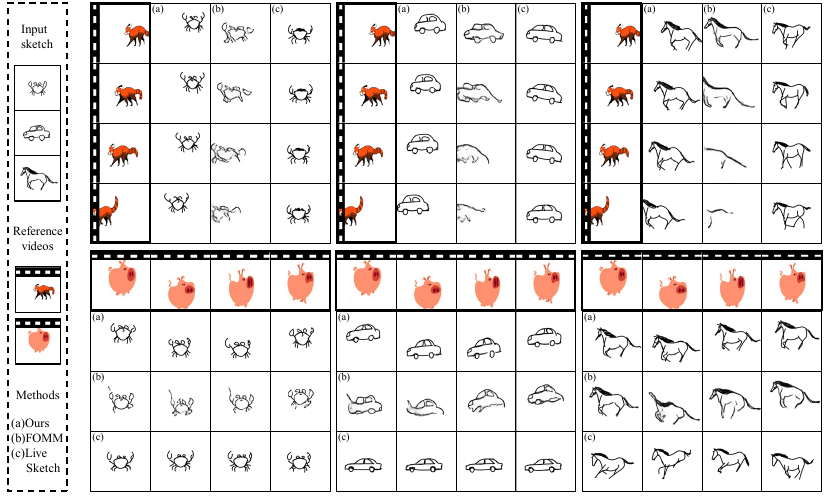}
  \caption{Qualitative comparisons with previous methods. FOMM captures good motion information but may cause deformation in the target sketch. Sketch videos generated using Live Sketch tend to be static or exhibit slight deformations. However, our approach effectively encodes motion information while avoiding changes in appearance.}

  \label{fig:comparison}
\end{figure*}
\vspace{-0.2cm}
\subsection{SketchAnimator}
Fig.~\ref{fig:pipline} illustrates our three-stage learning pipeline, which takes a user-provided static sketch in vector format and a driving video as input and produces the customized sketch video. We decouple this task into appearance learning, motion learning and video prior distillation.

\noindent\textbf{Appearance Learning}
The input sketch is firstly represented in the image format. Inspired by Dreambooth~\cite{ruiz2023dreambooth} and Textual Inversion~\cite{gal2023an}, we learn the appearance information of sketch by optimizing the attention layers with trainable LoRAs of the base T2V model, denoted as Appearance LoRAs (A-LoRAs). Also, we find that representing the sketch as a single subject by a semantic word (\emph{e.g.}, ``A horse."), denoted by $y_a$, serves as a better textual quires than the format of ``A photo of $*$." or ``A sketch drawing of $*$." Given that in this scenario, the input sketch image exhibits no temporal variations, we solely optimize spatial transformer layers in the 3D U-Net blocks following the objectives:
\vspace{-0.1cm}
\begin{equation}
    \mathcal{L}_{appearance} = \mathbf{E}_{z_0, y_a, t, \epsilon }\left[\left\| \epsilon - \epsilon_{W}\left(z_t, \tau_\theta(y_a), t\right)\right\|_2^2\right].
\label{eq:spatial_loss}
\end{equation}

\noindent\textbf{Motion Learning}
\label{sec:motion_learn}
To obtain the motion signal from the video, we decouple and reconstruct the video in both the temporal and spatial dimensions. As shown in Fig.~\ref{fig:pipline}, given a reference video and corresponding caption, we add LoRAs to attention blocks in spatial transformer and temporal transformer separately. For each training step, the A'-LoRAs are trained on a single frame randomly sampled from the training video to fit its appearance while ignoring its motion, based on spatial loss, which is reformulated as
\vspace{-0.1cm}
\begin{equation}
    \mathcal{L}_{spatial} = \mathbf{E}_{z_0, y_m, t, \epsilon, i\sim \mathcal{U}(0, \mathbf{\mathrm{F}})}\left[\left\| \epsilon - \epsilon_\theta\left(z_{t,i}, \tau_\theta(y_m), t\right)\right\|_2^2\right],
\label{eq:spatial_loss}
\end{equation}
where $F$ is the number of frames of the training data and the $z_t$, $i$ is the sampled frame from the latent code $z_t$ and the $y_m$ is the prompt depicting the video sequence.

For Motion LoRAs (M-LoRAs), the loss function is following the VDM loss as \eqref{eq:VDM_loss}. Therefore, the final objective
of our motion learning process is the combination of temporal loss and spatial loss as follows:
\vspace{-0.1cm}
\begin{equation}
    \mathcal{L}_{motion}=  \mathcal{L}_{spatial} + \mathcal{L}_{temporal}.
    \label{eq:motion_loss}
\end{equation}

\noindent\textbf{Video Prior Distillation}
Initially, the SVG sketch is consist of $N$ strokes for the first frame $P=\{p_1, ..., p_N\} \in \mathbf{R}^{N \times 4\times2}$. Each
stroke is represented by a two-dimensional \myBezier curve with four control point $p_i =\{p_i^j\}_{j=1}^4 = \{(x_i, y_i)^j\}_{j=1}^4 \in \mathbf{R}^{4 \times 2}$. Next, in accordance with the frame count of the reference video, the strokes are duplicated $F$ times for each frame initialization, which denoted by $S = \{P^f\}_{f=1}^F \in \mathbf{R}^{N \cdot F \times 4\times2}$. Then, all the strokes are fed into a differentiable rasterizer, $R\phi$, to acquire  raster frames.
Therefore, the input sequence in the latent space is noted as:
\vspace{-0.1cm}
\begin{equation}
    z_0 = E_\theta(R_\phi(S)),
\end{equation}
where $E_\theta$ is the pre-trained variational autoencoder.

Subsequently, we inject the A-LoRAs of sketch $\Delta W_A$ and M-LoRAs of reference video $\Delta W_M$ into the pre-trained video diffusion model as expected video prior, the final weight matrix of VDM is:
\vspace{-0.1cm}
\begin{equation}
    W' = W_{0} + \lambda_1 \Delta W_A  + \lambda_2 \Delta W_M,
\end{equation}
and then the \myBezier curve $S$ is updated using SDS loss until the structure, position and scale of sketch is changed as desired and we get final sketch curve $S'$ following:
\vspace{-0.1cm}
\begin{equation}\label{eq:sds_loss}
    \nabla_\phi \mathcal{L}_{SDS} = \left[ w(t)(\epsilon_{{W}'}(z_t,\tau_\theta(y)),t) \frac{\partial x}{\partial \phi} \right] ,
\end{equation}

\vspace{-0.1cm}
\begin{equation}
    \phi' \xleftarrow{\nabla_\phi \mathcal{L}_{SDS}} \phi ,
    \label{eq11}
\end{equation}
\vspace{-0.3cm}
\begin{equation}
    S' \xleftarrow{\phi'} S,
    \label{eq12}
\end{equation}
where $w(t)$ is a weighting function. The Eq.\eqref{eq11} and \eqref{eq12} are the brief backpropagation process for bridging the raster and vector domains by the differentiable rasterizer~\cite{li2020differentiable}.

\section{EXPERIMENTS}
\label{sec:experiment}
\subsection{Experimental Setup}

\noindent\textbf{Datasets}
The proposed method is evaluated on the popular motion transfer datasets
MGIF dataset~\cite{siarohin2019animating}. For each video, we pick 5 sketches from CLIPasso~\cite{vinker2022clipasso}, QuickDraw~\cite{cheema2012quickdraw} and SketchVOS~\cite{Yang_2023_BMVC}. In our experiments, all sketches are pre-processed into vector format and all the frames are pre-processed into 256 × 256 pixels. 
\begin{table}[t]
\footnotesize
    \centering
    \caption{Quantitative evaluations and ablation studies}
    \begin{tabular}{lccc} 
    \toprule
    \multirow{2}{*}{Method}  & Appearance & Motion  & Temporal \\
     & Alignment $\left(\uparrow\right)$ & Alignment$\left(\uparrow\right)$ & Consistency$\left(\uparrow\right)$ \\
    \midrule 
    FOMM & $0.824$& $0.209$ & $0.934$ \\
    Custom-A-Video & $0.689 $& $0.314$ & $0.879$ \\
    MotionDirector & $0.729 $& $0.398$ & $0.942$ \\
    DreamVideo & $0.743 $& $0.407$ & $0.951$ \\
    Live Sketch & $0.948 $ & $0.460$& $0.980 $ \\
    Ours & $\mathbf{0.955}$& $\mathbf{0.541}$ & $\mathbf{0.988}$ \\
    \midrule
    Ablation results & & & \\
    w/o A-LoRAs & $0.947 $ & $0.512 $ & $0.995$ \\
    w/o M-LoRAs & $0.952$ & $0.457 $ & $0.981$\\
    \bottomrule \\
\end{tabular}
\vspace{-0.2cm}
    \label{tab:all_quant}
\end{table}

\noindent\textbf{Implementation details}
We evaluate our approach primarily on Modelscope~\cite{wang2023modelscope}, a standard video diffusion model. We employ the Adam optimizer to train all the parameters on a single RTX 3090 GPU. For sketch appearance learning and video motion learning, the LoRAs are optimized by 500 iterations following~\cite{ruiz2023dreambooth}. During the video prior distillation stage,  the differentiable rasterizer are updated with learning rate at $2.0 \times 10^{-3}$. The hyperparameters $\lambda_1$ and $\lambda_2$
 of the appearance LoRAs and motion LoRAs are set to $0.5$ and $1$.

 \noindent\textbf{Evaluation metrics}
Assessing the quality of sketch animation poses a challenge due to the ground truth animations are not available.
We evaluate our approach with the following three metrics:
(1) \textit{Appearance Alignment} measures the average cosine similarity between CLIP~\cite{radford2021learning} image embeddings of all generated frames and the input sketch.
 (2) \textit{Motion Alignment} computes the generated video frames
and the inference prompt depicting reference video motion, in the form of X-CLIP Score~\cite{ni2022expanding}, a metric for general video recognition.
(3) \textit{Temporal Consistency}
 calculates CLIP image embeddings for all generated frames and present the average cosine similarity across all pairs of consecutive frames.

 \vspace{-0.3cm}
\subsection{Main results}
Table~\ref{tab:all_quant} and Fig.~\ref{fig:comparison} present comparisons with cutting-edge methods. Our method achieves the best results with a better trade-off between the visual quality and motion alignments among all methods. In particular, FOMM~\cite{siarohin2019first} leads to sketch deformation in details, especially when the object in the reference video differs significantly from the sketch in both shape and semantics. On the contrary, our approach disentangles a cleaner motion signal, leading to videos characterized by abundant variety. Live sketch~\cite{gal2023breathing} is capable of generating high-resolution images. However, the preservation of original structure is far from accurate and the animation results are closed to static as shown in 8th row. In contrast, our method is able to generate more creative sketch video (e.g., combining ``car" with ``jumping"). State-of-the-art customized video generation methods~\cite{zhao2023motiondirector,wei2023dreamvideo, ren2024customize} perform worse in both appearance preservation and motion translation correctness.

\begin{figure}[t]
  \centering
  \setlength{\abovecaptionskip}{-8pt}
  \includegraphics[width=1.0\linewidth]{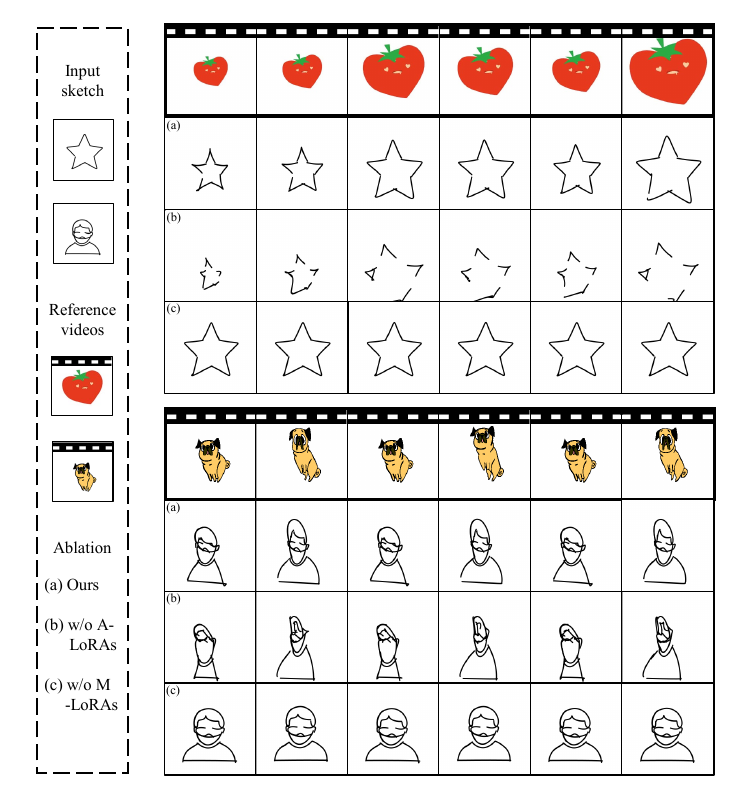}
  \caption{Visualization of ablation study. SketchAnimator faithfully retains both the identity of the subject and the pattern of its motion.}
  \vspace{-0.5cm}
  \label{fig:ablation}
\end{figure}

\vspace{-0.2cm}
\subsection{Ablation study}
We conduct an ablation study to demonstrate the necessity of each component. Specifically, we test two settings on the generator and report the results in the Table~\ref{tab:all_quant}: (1) w/o Appearance LoRAs (A-LoRAs), we observe a decrease in Appearance Alignment scores and the structure of the sketch undergoes deformation shown in the third row of Fig.~\ref{fig:ablation}. This suggests the effectiveness of learning sketch information. (2) w/o Motion LoRAs (M-LoRAs), we notice a significant decline in motion alignment scores and the sketch has transitioned to a static state illustrated in the fourth row of Fig.~\ref{fig:ablation}. This indicates that the M-LoRAs play a crucial role in predicting motion corresponding to the reference video.

\vspace{-0.2cm}
\subsection{Discussion}
\begin{figure}[!ht]
  \centering
  \setlength{\abovecaptionskip}{-8pt}  
  \setlength{\belowcaptionskip}{-9pt}
  \includegraphics[width=1.0\linewidth]{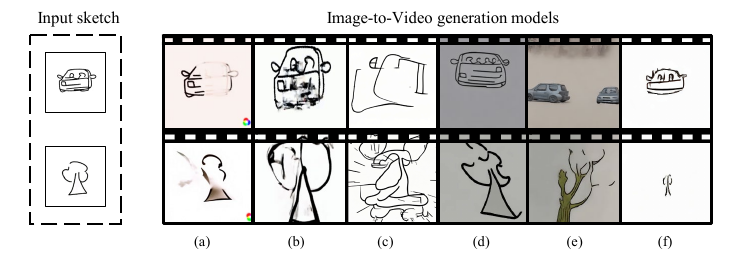}
  \caption{The results of Image-to-Video generation models. (a) Gen2, (b) SVD, (c) DynamiCrafter, (d) MotionDirector, (e) Custom-A-Video and (f) DreamVideo.}
  \label{fig:I2V}
\end{figure}

We further illustrate the results of advanced Image-to-Video generation models in Fig.~\ref{fig:I2V} including Gen2~\cite{RunwayGen2}, DynamiCrafter~\cite{xing2023dynamicrafter}, SVD~\cite{blattmann2023stable}, MotionDirector~\cite{zhao2023motiondirector}, Custom-A-Video~\cite{ren2024customize} and DreamVideo~\cite{wei2023dreamvideo}. All these methods show redundant information, and commonly fail to produce a sketch. In contrast, our method has significant advantages in these areas.

\vspace{-0.1cm}

\vspace{-0.6cm}
\section{Conclusion}
In this paper, we explore dynamic sketch generation customized by a reference video and construct a multi-stage
generation framework for better solving the challenges mentioned above. Extensive experiments demonstrate
that our approach achieved a high-quality and better motion alignment sketch animation generation compared to previous methods. This research opens up new possibilities for generating highly customized and creative animations tailored to individual users.


\end{document}